\documentclass[11pt]{article}

\usepackage[T1]{fontenc}
\usepackage[utf8]{inputenc}
\usepackage{lmodern}
\usepackage{comment}
\usepackage{float}

\usepackage{amsmath,amssymb}
\usepackage{graphicx}
\usepackage[hidelinks]{hyperref}

\usepackage{authblk}
\usepackage{orcidlink}

\usepackage[ruled, linesnumbered]{algorithm2e}
\usepackage{enumitem}
\usepackage{booktabs}
\usepackage[table]{xcolor} 
\usepackage{colortbl}
\usepackage{wasysym}
\usepackage{siunitx}
\usepackage{multirow}
\usepackage{fontawesome5} 
\newcommand{\rmark}[1]{\makebox[0pt][l]{\hspace{0.2em}#1}}

\definecolor{myLightGray}{gray}{0.75}
\newcommand{\lightmidrule}{%
  \arrayrulecolor{myLightGray}\specialrule{0.4pt}{0pt}{0pt}\arrayrulecolor{black}%
}

\definecolor{CMSAblue}{HTML}{4C78A8}
\newcommand{\psigA}{\textsuperscript{\textcolor{CMSAblue}{\textbf{*}}}}
\newcommand{\psigB}{\textsuperscript{\textcolor{CMSAblue}{\textbf{**}}}}
\newcommand{\psigC}{\textsuperscript{\textcolor{CMSAblue}{\textbf{***}}}}

\title{Construct, Merge, Solve \& Adapt with Reinforcement Learning for the min--max Multiple Traveling Salesman Problem}

\author[1,2]{Guillem Rodr\'iguez-Corominas\,\orcidlink{1234-5678-9012}}
\author[2]{Maria J. Blesa\,\orcidlink{1234-5678-9012}}
\author[1]{Christian Blum\,\orcidlink{1234-5678-9012}}

\affil[1]{Artificial Intelligence Research Institute (IIIA-CSIC), Bellaterra, Catalunya, Spain\\
\texttt{grodriguez@iiia.csic.es}, \texttt{christian.blum@iiia.csic.es}}
\affil[2]{Universitat Polit\`ecnica de Catalunya (UPC), Barcelona, Catalunya, Spain\\
\texttt{maria.j.blesa@upc.edu}}

\date{} 

\begin{document}
\maketitle

\begin{abstract}
The Multiple Traveling Salesman Problem (mTSP) extends the Traveling Salesman Problem to $m$ tours that start and end at a common depot and jointly visit all customers exactly once. In the min--max variant, the objective is to minimize the longest tour, reflecting workload balance. We propose a hybrid approach, \textit{Construct, Merge, Solve \& Adapt} with Reinforcement Learning (\textsc{RL-CMSA}), for the symmetric single-depot min--max mTSP. The method iteratively constructs diverse solutions using probabilistic clustering guided by learned pairwise $q$-values, merges routes into a compact pool, solves a restricted set-covering MILP, and refines solutions via inter-route remove, shift, and swap moves. The $q$-values are updated by reinforcing city-pair co-occurrences in high-quality solutions, while the pool is adapted through ageing and pruning. This combination of exact optimization and reinforcement-guided construction balances exploration and exploitation. Computational results on random and TSPLIB instances show that \textsc{RL-CMSA} consistently finds (near-)best solutions and outperforms a state-of-the-art hybrid genetic algorithm under comparable time limits, especially as instance size and the number of salesmen increase.
\end{abstract}


\section{Introduction}
The Multiple Traveling Salesman Problem (mTSP) generalizes the TSP to $m$ tours that start and end at a common depot and collectively visit each customer exactly once. It can be viewed as a simplified vehicle routing problem (VRP) without vehicle capacities or time windows \cite{Tang2000,Park2001}. Two objectives are usually considered: (i) minimizing total traversal cost (the \emph{min-sum mTSP}) and (ii) minimizing the length of the longest tour among the $m$ tours (the \emph{min--max mTSP}). The min--max criterion is particularly relevant when balancing workload or service time is important, as in last-mile delivery with identical vehicles, coordinated multi-robot patrolling, UAV sortie planning, or technician routing, where fairness and service-level constraints are crucial~\cite{Wang2017}. This paper focuses on the single-depot min--max mTSP on symmetric graphs, a setting that captures many practical applications while remaining computationally challenging.

Formally, let $G=(V,E)$ be a complete undirected graph with $V=\{0,1,\dots,n\}$, where $0$ denotes the depot and $D_{ij}$ is the metric cost of edge $(i,j)$. The task is to construct $m$ closed routes $\{\mathcal{R}_1,\dots,\mathcal{R}_m\}$ that (i) are node-disjoint except for the depot, (ii) cover $\{1,\dots,n\}$ exactly once, and (iii) minimize $\max_{k=1,\dots,m} \mathrm{len}(\mathcal{R}_k)$, where $\mathrm{len}(\cdot)$ is the tour cost. The problem is NP-hard even for metric instances by reduction from the TSP and balanced partitioning arguments. Consequently, exact algorithms scale only to modest sizes, while heuristics and metaheuristics dominate large-instance practice.

In recent years, hybrid metaheuristics that combine heuristic search with exact optimization components have shown strong potential for tackling difficult combinatorial problems. Along these lines, we build on the \textit{Construct, Merge, Solve \& Adapt} (\textsc{CMSA}) framework, which is itself a hybrid metaheuristic integrating constructive solution generation with an exact solving phase. In particular, we consider its Reinforcement Learning extension, \textsc{RL-CMSA} \cite{Reixach2024}, which augments standard \textsc{CMSA} with a Q-learning mechanism to guide and adapt key decisions during the search. We argue that this additional learning-driven adaptation can be especially beneficial for the problem studied in this paper.

\section{Related Work}

Methodologically, mTSP algorithms largely follow the standard routing taxonomy: \emph{exact} approaches (e.g., branch-and-cut on arc/flow models with subtour elimination, or set-partitioning with branch-and-price) and \emph{heuristics/metaheuristics} that trade optimality guarantees for scalability \cite{Soylu2015,Venkatesh2015}. Within the latter, two recurring design patterns are \emph{cluster-first route-second} (assign customers to $m$ salesmen via balanced partitioning methods and then solve $m$ TSPs) and \emph{route-first cluster-second} (build a giant tour and optimally \emph{split} it into $m$ routes via shortest-path or dynamic-programming procedures) \cite{Mahmoudinazlou2024}. Modern hybrids strengthen these templates with powerful local search (e.g., 2-opt/Or-opt/$k$-exchange), tailored recombination, and diversity control, often yielding state-of-the-art performance in both min-sum and min--max settings.

\begin{figure}[h!]
  \centering
  {\includegraphics[width=\textwidth]{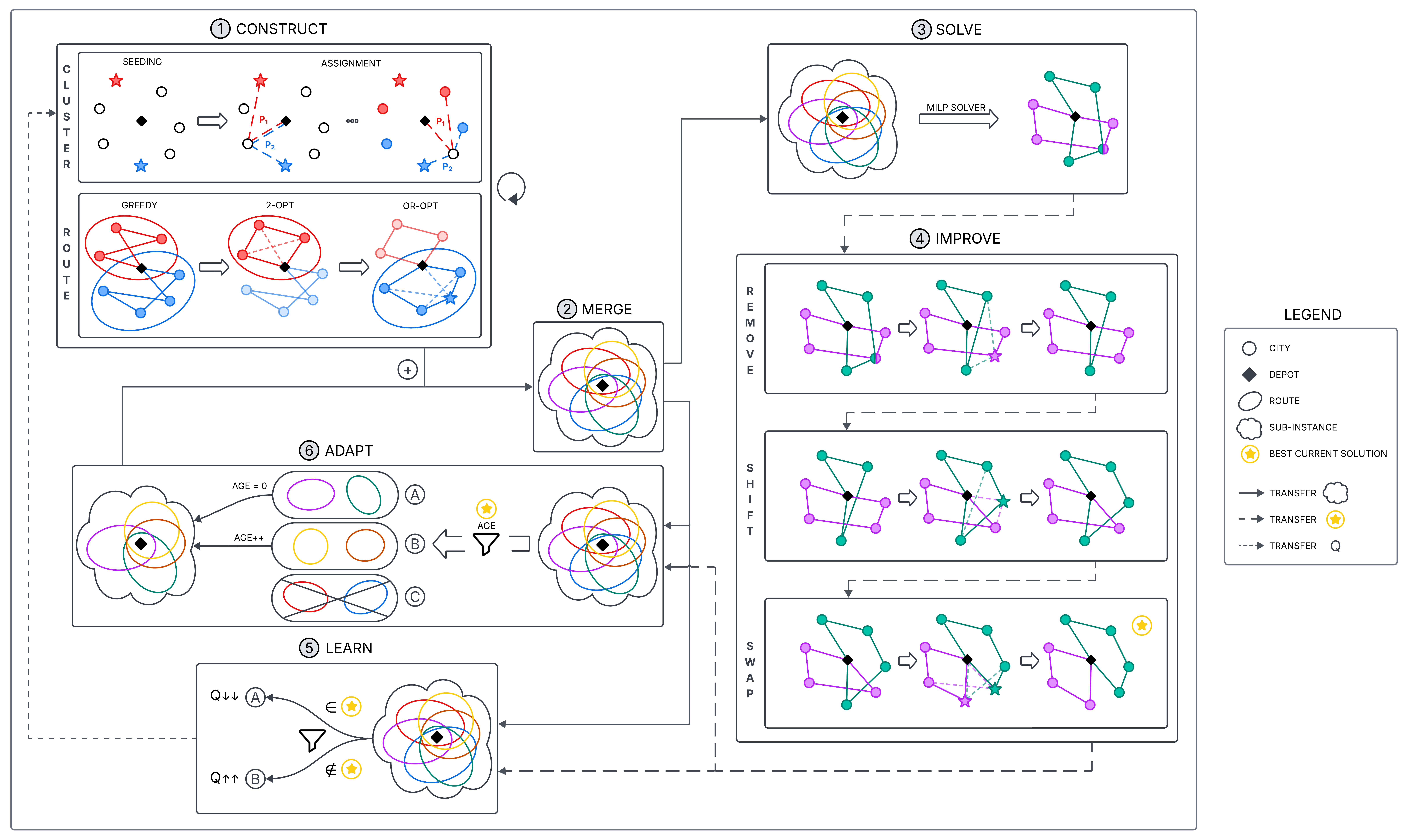}}
  \caption{\textsc{RL-CMSA}: Schematic Algorithm Overview}
  \label{fig:cmsa_tsp}
\end{figure}

Genetic Algorithms are particularly prominent in the mTSP literature, evolving either explicit multi-route representations or TSP-like permutations with route delimiters, combined with selection and diversity mechanisms and typically hybridized with local improvement \cite{JunLi2013}. For the min--max mTSP, a strong representative is the Hybrid Genetic Algorithm (HGA) of~\cite{Mahmoudinazlou2024}, which encodes solutions as TSP sequences, evaluates individuals through a DP-based \emph{Split} tailored to the min--max objective, and combines a similarity-aware crossover with both inter- and intra-route local search (e.g., 1-shift/1-swap/Or-opt/2-opt), plus intersection handling. When tested on four standard benchmark suites---including, for example, TSPLIB instances and random instances---the \textsc{HGA} achieves state-of-the-art average performance under comparable cutoffs and improves best-known results on a substantial subset of instances, making it a leading baseline for our setting. Therefore, in this paper, we study the min–max mTSP under the same modeling assumptions and use the \textsc{HGA} as a strong baseline for comparison.

Beyond GAs, mTSP variants have been tackled by Ant Colony Optimization \cite{Yousefikhoshbakh2012}, Artificial Bee Colony \cite{Tong2025}, Invasive Weed Optimization \cite{Venkatesh2015}, General Variable Neighborhood Search \cite{Soylu2015}, and Memetic Algorithms \cite{Wang2017, He2023}, to name a few. For  the min--max variant, competitive examples include the Hybrid Search with Neighborhood Reduction (HSNR) \cite{He2022}, which alternates tabu-based inter-tour moves with EAX-based intra-tour improvement; and the Iterated Two-Stage Heuristic (ITSHA) \cite{Zheng2022}, combining randomized or clustering-based initialization with variable neighborhood improvement (2-opt/insert/swap).

Finally, Reinforcement Learning (LR) and GNN-based methods have emerged as promising alternatives. Notable examples include multi-agent RL frameworks (e.g., ScheduleNet) \cite{Park2021} and Neuro--Cross Exchange (NCE), which learns to activate a cross-exchange neighborhood \cite{Kim2022}. These methods can produce competitive solutions quickly, though classical/hybrid metaheuristics still dominate many standard benchmarks under typical time budgets.

\section{The Proposed Algorithm}

A schematic overview of our \textsc{RL-CMSA} algorithm is given in Figure~\ref{fig:cmsa_tsp}. The algorithm iterates through six phases—\emph{Construct}, \emph{Merge}, \emph{Solve}, \emph{Improve}, \emph{Learn}, and \emph{Adapt}—until the stopping criterion is met (here, a time limit).  In the following sections, all algorithm components are explained in detail.

\subsection{Construct}

In the first phase of the algorithm, we generate $n_{\text{solutions}}$ candidate solutions through a probabilistic construction process. Each construction is divided into two stages: \textsc{Cluster} and \textsc{Route}. \\

\noindent \textsc{Cluster} stage:  The goal of this stage is to partition the set of cities $V$ (excluding the depot) into $m$ clusters, one for each vehicle. The intuition is that cities that are likely to belong to the same route should be grouped together. Algorithm~\ref{algo:cluster} provides pseudocode for this procedure.
    
The clustering process begins by selecting $m$ \emph{centers}, which serve as the initial representatives for each cluster (see \texttt{Seeding}). These centers are chosen using a $k$-means++ style seeding procedure, biased by the $q$-values of city pairs (see lines 1-12 of Algorithm~\ref{algo:cluster}):
    \begin{enumerate}[leftmargin=*]
        \item The first center is sampled from the non-depot cities with probability proportional to the squared distance from the depot.
        \item Each subsequent center is chosen using a roulette-wheel selection, where the probability of selecting a city $i$ depends on its squared distance to the nearest already-chosen center, adjusted by the corresponding $q$-value.
    \end{enumerate}
    This ensures that centers are both well separated and more likely to belong to different optimal routes. Figure~\ref{fig:cmsa_tsp} illustrates this process, where the chosen centers are marked with a star.

    After the centers are chosen, the remaining cities are assigned to clusters (\texttt{Assignment}). The so far unassigned cities from $U$ are then assigned one by one to the clusters using a weighted assignment mechanism (see lines 13-32 of Algorithm~\ref{algo:cluster}). Cities with smaller angular distance (with respect to the depot) to a center ($\mathrm{angdist}(\theta_u,\theta_{c_j})$) are more likely to be processed earlier. For each candidate city $u$, we compute the approximate assignment cost into each cluster $C_j$, defined as the distance to its two closest points in that cluster, including the depot when applicable (function \textsc{twoClosestPoints}()). When scoring the assignment of a city \(u\) into a cluster \(C_j\) (see line~26), we scale the cost by two additional factors: (i) the \emph{mean \(Q\)-compatibility} between \(u\) and all current members of \(C_j\), excluding the depot (see line~24), and (ii) a \emph{running estimate} \(r_{u,j}\) of how the max length would increase if the city is added to this cluster (see line~25).\footnote{Note that $d_{u,j}$ and $ L^{\text{approx}}_{j}$ are proxies computed from anchor distances and are used only to guide clustering. Actual tours are constructed in the subsequent \textsc{Route} stage.} Note that a small \(\varepsilon>0\) is added for numerical stability.

    Once the costs are computed, with probability $d_{\text{rate}}^{\text{construct}}$, the city is assigned to the cluster with the lowest score. Ties are broken uniformly at random using reservoir sampling. Otherwise, the cluster is chosen with probabilities $p_j \;\propto\; \frac{1}{s_j}$. The factor \(L^{\text{approx}}_j\) is updated incrementally after each insertion using the corresponding two-anchor insertion cost, which implicitly encourages \emph{load balancing} by penalizing already long provisional routes.  
    
    This procedure is repeated until all cities have been assigned to clusters. \\

    \begin{algorithm}[H]
    \caption{Clustering algorithm biased by $q$-values}
    \label{algo:cluster}
    \KwIn{$D$: pairwise distances; $Q$: $q$-values; $m$: \# routes}
    \KwOut{$\mathcal{R}=\{C_1,\ldots,C_m\}$: $m$ clusters}

    \BlankLine
    \textbf{Phase 1:} \texttt{Seeding} \\
    Initialize $U \gets V\setminus\{0\}$ \tcp*[r]{Non-depot cities}
    Initialize $w_i \gets D_{0,i}^2$ for all $i \in U$\;

    \For{$j=1$ \KwTo $m$}{
        Sample center $c_j \in U$ with probability $\propto w_i$\;
        $U \gets U \setminus \{c_j\}$\;
        Initialize cluster $C_j \gets \{\,0,\,c_j\,\}$\;
        Initialize $L^{\text{approx}}_j \gets 2\cdot D_{0,c_j}$\;
        \ForEach{$i \in U$}{
            $w_i \gets \min\{\,w_i,\; D_{i,c_j}^2 \cdot Q_{i,c_j}^2\,\}$\;
        }
    }

    \BlankLine
    \textbf{Phase 2:} \texttt{Assignment} \\
    \ForEach{$u \in U$}{   
        $\delta_u \gets \min_{j=1,\ldots,m}\mathrm{angdist}(\theta_u,\theta_{c_j})$\;
    }

    \While{$U \neq \emptyset $}{
        Sample city $u \in U$ with probability $\propto 1/\delta_u$\;
        $U \gets U \setminus \{u\}$\;

        $L_{\max}^{\text{approx}} \gets \max_{j=1,\ldots,m} L^{\text{approx}}_j$\;
        \ForEach{$j=1,\ldots,m$}{
            $\{a,b\} \gets \textsc{twoClosestPoints}(u, C_j, D)$\;
            $d_{u,j} \gets D_{u,a}+D_{u,b}-D_{a,b}$\;
            $q^{\text{mean}}_{u,j} \gets \dfrac{1}{|C_j \setminus \{0\}|}\sum_{v \in C_j,\ v \neq 0} Q_{u,v}$\;
           $r_{u,j} \gets \max\!\bigl(L_{\max}^{\text{approx}},\,L^{\text{approx}}_j + d_{u,j}\bigr)/L_{\max}^{\text{approx}}$\;
            $s_j \gets (d_{u,j} + \varepsilon) \cdot r_{u,j} \cdot    q^{\text{mean}}_{u,j}$\;
        }

        With probability $d_{\text{rate}}^{\text{construct}}$: choose $j^\star \in \arg\min_j s_j$\;
        Otherwise: sample $j^\star$ with probability $p_j \propto 1/s_j$\;
        $C_{j^\star} \gets C_{j^\star} \cup \{u\}$\;
        $L^{\text{approx}}_{j^\star} \gets L^{\text{approx}}_{j^\star} + d_{u,j^\star}$\;
    }

    \Return{$\mathcal{R}=\{C_1,\ldots,C_m\}$}\;
    \end{algorithm}

\noindent \textsc{Route} stage: Once the clustering is complete, we build one route per cluster using a fast greedy insertion heuristic. Starting from the depot, we construct a tour for each cluster using best-insertion into the partial route. The resulting tour is then improved by applying intra-route local search: first a 2-opt procedure and then an Or-opt procedure restricted to relocating sequences of length 1 and 2. Both local searches scan candidate moves in random order and apply first improvement, to avoid systematic bias. This produces high-quality cluster tours at low computational cost. 

After all routes of a solution have been constructed, we apply the inter-route improvement operators described in Section~\ref{sec:improve}. To reduce computational cost, we restrict the neighborhood to moves involving the longest route only, rather than considering all route pairs. \\

By repeating the above procedure $n_{\text{solutions}}$ times, \textsc{RL-CMSA} generates a set of $n_{\text{solutions}}$ solutions to the problem.

\subsection{Merge}

Next, in the \emph{Merge} step of \textsc{RL-CMSA} (see the center of Figure~\ref{fig:cmsa_tsp}), the $n_{\text{solutions}}\times m$ routes generated in the \emph{Construct} step are added to the pool $\mathcal{R}_{\text{cand}}$ of candidate routes. Each route stored in $\mathcal{R}_{\text{cand}}$ is assigned an \textit{age} value, which is initialized to $0$ upon insertion.\footnote{This will be important for the \emph{Adapt} step explained in Section~\ref{sec:adapt}.}

Since the construction process may generate multiple routes visiting the same set of non-depot cities, we keep only one representative per visited-city set, namely the shortest one. To do so, we compute a canonical signature for each route by sorting its non-depot cities. This signature is then used for hashing and equality checks when inserting into $\mathcal{R}_{\text{cand}}$.

Finally, we prune $\mathcal{R}_{\text{cand}}$ by discarding any route whose length exceeds the current incumbent max-route length. Such routes will never be selected by the solver (as explained in the following section), and keeping them would only bias the subsequent learning step.

\subsection{Solve}

After the \emph{Merge} step, \textsc{RL-CMSA} conducts the \emph{Solve} step (top right of Figure~\ref{fig:cmsa_tsp}) to determine the best solution that can be assembled from the candidate pool $\mathcal{R}_{\text{cand}}$. We formulate a \emph{set-covering} MILP that selects exactly $m$ routes while ensuring that every non-depot city is covered at least once, and minimizes the length of the longest selected route. The resulting subproblem is solved with \texttt{CPLEX}, a state-of-the-art MILP solver.

More formally, let $x_r \in \{0,1\}$ be a binary variable indicating whether route $r \in \mathcal{R}_{\text{cand}}$ is selected, and let $z$ be an upper bound on the maximum length among selected routes. The model is:
\begin{align}
  \min \quad & z \nonumber \\
  \text{s.t.}\quad
  & z \ge \ell_r - M(1-x_r) && \forall r \in \mathcal{R}_{\text{cand}} \nonumber \\
  & \sum_{r \in \mathcal{R}_{\text{cand}}} x_r = m \nonumber \\
  & \sum_{r \ni i} x_r \ge 1 && \forall i \in V \setminus \{0\} \nonumber
\end{align}
where $\ell_r$ denotes the length of route $r$ and $M$ is a sufficiently large constant. Note that not forcing each city to be covered by only one route allows for greater flexibility. However, as the selected routes may overlap, the output $\mathcal{R}_{\text{best}}$ is not necessarily a valid mTSP solution yet, and duplicates are resolved in the subsequent \emph{Improve} step. Note that we can impose $z$ to be bounded by the current incumbent best solution, which justifies pruning such routes from $\mathcal{R}_{\text{cand}}$ in the previous step.

\subsection{Improve}
\label{sec:improve}

In the \emph{Improve} step of \textsc{RL-CMSA} (see bottom right in Figure~\ref{fig:cmsa_tsp}), $\mathcal{R}_{\text{best}}$ is refined by applying a series of local improvement operations to its routes. The primary objective is to minimize the length of the longest route in $\mathcal{R}_{\text{best}}$, henceforth denoted by $z$.

\subsubsection{Remove}
This step turns $\mathcal{R}_{\text{best}}$ into a feasible solution (if necessary) by removing non-depot cities occurring in more than one route in the following way.

For each duplicated city, we evaluate the potential decrease in route length that would result from removing the city from a given route by directly connecting its predecessor and successor. Let $u$ denote the city to be removed, and let $a$ and $b$ be its predecessor and successor in the route, respectively. The \emph{removal improvement} $\Delta_{\text{remove}}$ is defined as 
\[
\Delta_{\text{remove}}(u) \;=\; D_{au} + D_{ub} - D_{ab} \quad.
\]
Once these improvement values have been computed for all duplicated cities across all routes, we proceed as follows: with probability $d_{\text{rate}}^{\text{improve}}$, we remove the city yielding the highest removal improvement; otherwise, we select a city to remove using a roulette-wheel mechanism, with probabilities proportional to their respective $\Delta_{\text{remove}}$ values. These values are scaled by the respective current route length to prioritize deletions from longer routes. This process is repeated iteratively until every non-depot city appears in exactly one route.

\subsubsection{\textsc{Shift}: cross-route relocation (1-move)}

Aiming for further improvement, we attempt to \emph{shift} a single city $u$ across routes, that is, to relocate it from its current source route into some position within a different target route of $\mathcal{R}_{\text{best}}$. The goal of this operation is to reduce the total route length by moving cities that are poorly placed, i.e., far from their neighboring cities, into routes that are closer. For all eligible cities and insertion positions, the following values are computed:
\[
\Delta^{-}_{\text{src}}(u) = D_{a u} + D_{u b} - D_{a b} \quad \text{(removal gain)},
\]
\[
\Delta^{+}_{\text{tgt}}(p,u) = D_{c u} + D_{u d} - D_{c d} \quad \text{(insertion cost)},
\]
where $a$ and $b$ are the predecessor and successor of $u$ in its source route, and $(c,d)$ denotes the edge at the target insertion position $p$. Therefore, a relocation move has a net value given by:
\[
\Delta_{\text{shift}} = \Delta^{-}_{\text{src}}(u) - \Delta^{+}_{\text{tgt}}(p,u) \quad,
\]
and is considered \emph{admissible} only if the maximum route length after applying the move does not exceed $z$. Note that, with $\Delta_{\text{shift}} > 0$, the move strictly improves the secondary objective (total length), so it is unambiguously beneficial.

Since the primary objective is to minimize the length of the longest route in $\mathcal{R}_{\text{best}}$, we also allow admissible moves with $\Delta_{\text{shift}} < 0$: these may slightly increase the total length but still improve the main objective. To balance exploitation and exploration, we select among admissible moves as follows: with probability $d_{\text{rate}}^{\text{improve}}$ we select the best move; otherwise we sample a move via roulette–wheel with probabilities proportional to $\exp\!\bigl(\frac{\Delta_{\text{shift}}}{z}\bigr)$. This weighting favors moves that reduce both $z$ and the total length ($\Delta_{\text{shift}} > 0$), while still giving a small but nonzero chance to moves with $\Delta_{\text{shift}} < 0$ when they reduce $z$, our primary objective. After applying the chosen move, only the affected $\Delta$ values around the modified positions are incrementally updated. This repeats iteratively until no admissible relocation remains.

\subsubsection{\textsc{Swap}: cross-route exchange (1--1 swap)}

In this phase, we attempt to \emph{swap} two cities $u$ and $v$ belonging to different routes from $\mathcal{R}_{\text{best}}$, $r_{\text{src}}$ and $r_{\text{tgt}}$ (${\text{src}} \neq {\text{tgt}}$), respectively. The objective is to further reduce route lengths by exchanging cities that may be better suited to each other’s routes. The incremental cost of such a swap is evaluated locally around the affected positions in each route. Let $a$ and $b$ denote the predecessor and successor of $u$ in $r_{\text{src}}$, and $c$ and $d$ those of $v$ in $r_{\text{tgt}}$. The respective local cost variations are computed as:
\[
\Delta_{\text{src}}(u \leftrightarrow v) = (D_{a u} + D_{u b}) - (D_{a v} + D_{v b}),
\]
\[
\Delta_{\text{tgt}}(u \leftrightarrow v) = (D_{c v} + D_{v d}) - (D_{c u} + D_{u d}).
\]
The total gain from performing the swap is then given by:
\[
\Delta_{\text{swap}} = \Delta_{\text{src}}(u \leftrightarrow v) + \Delta_{\text{tgt}}(u \leftrightarrow v).
\]
Similarly to the shift move, a swap is considered \emph{admissible} only if the maximum route length after applying the swap does not exceed $z$. Furthermore, we also allow moves with $\Delta_{\text{swap}} \leq 0$ if they improve the maximum route length $z$. Again, we select the swap with the maximum $\Delta_{\text{swap}}$ with probability $d_{\text{rate}}^{\text{improve}}$. Otherwise, we sample a move using a roulette wheel with probabilities proportional to $\exp\!\bigl(\frac{\Delta_{\text{swap}}}{z} \bigr)$. After applying the swap, only the cost entries in the neighborhoods of the modified positions are incrementally updated. The process repeats iteratively until no admissible improving swap moves remain. \\

The final solution $\mathcal{R}_{\text{best}}$ is the current best solution of \textsc{RL-CMSA}. This solution is marked in Figure~\ref{fig:cmsa_tsp} with a golden, encircled, asterisk. It serves as input to the  \emph{Learn} and \emph{Adapt} steps, as explained below.


\subsection{Learn}

We estimate two symmetric co-occurrence counts over unordered pairs $\{i,j\}$ of non-depot cities:
\begin{align}
S_{\text{cand}}(i,j) & =\#\{\text{routes in $\mathcal{R}_{\text{cand}}$ containing both } i,j\} \nonumber \\
S_{\text{best}}(i,j) & =\#\{\text{routes in $\mathcal{R}_{\text{best}}$ containing both } i,j\} \nonumber
\end{align}
In other words, $S_{\text{best}}(i,j)$ evaluates to either zero or one. For pairs that appear in the candidate pool ($S_{\text{cand}}(i,j)>0$), the following update of the corresponding $q$-values is conducted:
\[
Q_{ij}\leftarrow
\begin{cases}
Q_{ij} - l_{\text{rate}}\cdot Q_{ij}, & \text{if } S_{\text{best}}(i,j)>0 \quad\text{(reinforce)}\\[2pt]
Q_{ij} + l_{\text{rate}}\cdot (1-Q_{ij}), & \text{if } S_{\text{best}}(i,j)=0 \quad\text{(discourage)}.
\end{cases}
\]
Thus, beneficial pairs drift toward $0$ (more likely to end up in the same cluster in the \emph{Construct} step), while unhelpful pairs drift toward $1$. Moreover, a convergence proxy 
\(
\frac{2}{(n-1)(n-2)}\sum_{i<j}\!\big|\tfrac{1}{2}-Q_{ij}\big|
\)
is monitored over a short time window. If it stagnates, we reset all $q$-values to 0.5 and clear the sub-instance.

\subsection{Adapt}
\label{sec:adapt}

In the \emph{Adapt} step of \textsc{RL-CMSA} (see center left of Figure~\ref{fig:cmsa_tsp}), the sub-instance $\mathcal{R}_{\text{cand}}$ is updated using an age-based policy. Routes in $\mathcal{R}_{\text{best}}$ that are not yet in $\mathcal{R}_{\text{cand}}$ are inserted with age $0$. Conversely, the age of each route in $\mathcal{R}_{\text{cand}}$ that is not part of $\mathcal{R}_{\text{best}}$ is increased by one. Finally, any route whose age reaches the threshold $\text{age}_{\max}$ is removed. This policy bounds the size of $\mathcal{R}_{\text{cand}}$, promotes turnover, and keeps the pool compact and up-to-date with the evolving reinforcement signal.


\section{Experimental Evaluation}

The \textsc{RL-CMSA} algorithm described in the previous section stops at a user-defined computation time limit. It returns the best solution found. To assess the behaviour and applicability of our approach, we compare it with the state-of-the-art \textsc{HGA} algorithm from~\cite{Mahmoudinazlou2024}. All experiments were performed on the \textsc{Ars Magna} computing cluster at IIIA-CSIC (Intel Xeon 2.2\,GHz, 10 cores, 92\,GB RAM). For each instance, both algorithms were executed with a time limit of $n$ seconds (the number of cities).

\subsection{Benchmark Instances}

We considered two types of benchmark instances. First, and most importantly, we used randomly generated instances. Random instances are the most-used ones in the mTSP literature; however, for the min--max mTSP, they may become unintentionally easy when one or a few cities lie far from the depot. In such cases---especially as the number of vehicles increases---the optimal solution often consists of a few-city route to visit the distant city (or cities), while the remaining routes cover only nearby cities. To mitigate this effect and obtain more challenging instances, we generated random instances as follows: city locations are sampled uniformly at random inside a circle of radius $1$, placing the depot at the centre. This construction increases the likelihood that several cities lie near the boundary and thus at comparable distances from the depot. We generated 20 random instances for each $n \in \{50, 100, 200\}$. All these instances will be solved for $m \in \{\frac{1}{100}n, \frac{5}{100}n, \frac{10}{100}n, \frac{15}{100}n\}$. In other words, the number of salesmen considered ($m$) depends on the number of cities ($n$) and is calculated as $1\%$, $5\%$, $10\%$ or $15\%$ of the number of cities (rounded).

Second, we also considered instances from \texttt{TSPLIB}, which are commonly used in related work. In particular, we used \texttt{eil51}, \texttt{berlin52}, \texttt{eil76}, and \texttt{rat99}. These instances allow us to evaluate our approach on structured, well-known benchmarks and to test its behaviour across different problem characteristics.

\subsection{Algorithm Tuning}

To ensure good performance of our algorithm, a scientific parameter tuning was performed using \textsc{i-Race}, an automatic algorithm configuration tool. In particular, parameter tuning was performed independently for each considered value of $m$ (see above). For each $m$, a training set of five additional random instances with $200$ nodes (the largest size considered) was generated.

\begin{table}[H]
\centering
\caption{Summary of the parameters of \textsc{RL-CMSA} with their respective domain and the tuning results.}
\label{tab:tuning}
\resizebox{\columnwidth}{!}{%
\begin{tabular}[c]{ll|cccc} 
\hline
\multirow{2}{*}{\textbf{Parameters}} & \multirow{2}{*}{\textbf{Considered Domain}} & \multicolumn{4}{c}{\textbf{Tuning results}} \\
 & & $m=1\%$ & $m=5\%$ & $m=10\%$ & $m=15\%$ \\
\hline
$n_{\text{solutions}}$ & $\{2, 3, ..., 19, 20\}$ & $19$ & $17$ & $13$ & $17$ \\

$d_{\text{rate}}^{\text{construct}}$ & $\{0.00,0.01,...,0.89, 0.90\}$ & $0.87$ & $0.83$ & $0.66$ & $0.86$ \\

$d_{\text{rate}}^{\text{improve}}$ & $\{0.00,0.01,...,0.99, 1.00\}$ & $0.96$ & $0.97$ & $0.98$ & $0.93$ \\ 

$l_{\text{rate}}$ & $\{0.10,0.11,...,0.49, 0.50\}$ & $0.31$ & $0.26$ & $0.45$  & $0.20$ \\

$\text{age}_{\text{max}}$ & $\{2, 3, ..., 19, 20\}$ & $12$ & $13$ & $2$ & $15$ \\ \hline
\end{tabular}

}
\end{table}

Table~\ref{tab:tuning} reports the parameter domains explored by \textsc{i-Race} and the best configurations obtained. Overall, a clear pattern emerges for $m \in \{1\%, 5\%, 15\%\}$: both $n_{\text{solutions}}$ and $\text{age}_{\text{max}}$ take relatively large values, suggesting that maintaining a larger sub-instance is beneficial. In addition, $l_{\text{rate}}$ is set to a medium value, indicating that the algorithm tends to learn more slowly. Concerning the determinism rates, $d_{\text{rate}}^{\text{construct}}$ and $d_{\text{rate}}^{\text{improve}}$ are high (the latter is close to $1$), which implies that newly constructed solutions are kept close to the current best solution. This behaviour is consistent with an intensification strategy: the algorithm focuses the search in the neighbourhood of the best-so-far solution, and retaining a large sub-instance helps include routes similar to the incumbent, enabling a more thorough local exploration.

In contrast, the configuration found for $m=10$ differs substantially. Here, $n_{\text{solutions}}$ takes a more moderate value and $\text{age}_{\text{max}}$ is very small, which leads to smaller sub-instances and a shorter retention of seemingly bad routes. Combined with a lower value of $d_{\text{rate}}^{\text{construct}}$, this indicates a stronger emphasis on diversification: the algorithm generates more diverse candidates and keeps only the highest-quality ones. Moreover, the relatively high $l_{\text{rate}}$ suggests faster convergence.

For the \textsc{HGA} algorithm, we used the parameter values reported by the authors in~\cite{Mahmoudinazlou2024}. These values were tuned on a diverse set of instances, including random instances and the \texttt{TSPLIB} instances considered in this study. 

\subsection{Results}

Tables~\ref{tab:summary_m_1}(a)-(d) summarize the comparative results of \textsc{RL-CMSA} and \textsc{HGA} for each value of $m$ on the random instances. Each row corresponds to one instance (identified by \texttt{id}). For each problem size $n \in \{50,100,200\}$, the tables report, separately for \textsc{RL-CMSA} and HGA, the average objective value over $40$ independent runs (\texttt{mean}), the number of runs in which the method achieved the best value among the two compared algorithms (\texttt{\#b}), and the average time in seconds for finding the best solution of a run (\texttt{time}). In addition, the last column of each table block (regarding $n$ and $m$) reports the best objective value obtained on that instance across all runs and both algorithms (\texttt{best}). Boldface highlights the best \texttt{mean} value between the two algorithms for the corresponding instance and size, while the shaded cells provide a visual cue of which method performs better (blue for \textsc{RL-CMSA}, red for \textsc{HGA}, and grey for ties). The arrow and bolt symbols indicate, respectively, which algorithm attains a higher best-runs count and which one is faster for the corresponding instance, following the same color pattern.

\begin{table}[H]
\centering
\caption{Summary of the obtained results on random instances.}
\label{tab:summary_m_1}
\resizebox{\textwidth}{!}{%
\begin{tabular}{rcrrccrrcccrrccrrcccrrccrrc}
\toprule
\texttt{n} & \multicolumn{8}{c}{$50$} &  & \multicolumn{8}{c}{$100$} &  & \multicolumn{8}{c}{$200$} \\
\cmidrule(lr){2-9} \cmidrule(lr){11-18} \cmidrule(lr){20-27}
\textit{} & \multicolumn{3}{c}{\textcolor[HTML]{D95F5F}{{\bfseries\scshape HGA}}} & \multicolumn{1}{c}{} & \multicolumn{3}{c}{\textcolor[HTML]{4C78A8}{{\bfseries\scshape RL-CMSA}}} & \multicolumn{1}{c}{} &  & \multicolumn{3}{c}{\textcolor[HTML]{D95F5F}{{\bfseries\scshape HGA}}} & \multicolumn{1}{c}{} & \multicolumn{3}{c}{\textcolor[HTML]{4C78A8}{{\bfseries\scshape RL-CMSA}}} & \multicolumn{1}{c}{} &  & \multicolumn{3}{c}{\textcolor[HTML]{D95F5F}{{\bfseries\scshape HGA}}} & \multicolumn{1}{c}{} & \multicolumn{3}{c}{\textcolor[HTML]{4C78A8}{{\bfseries\scshape RL-CMSA}}} & \multicolumn{1}{c}{} \\
\cmidrule(lr){2-4} \cmidrule(lr){6-8} \cmidrule(lr){11-13} \cmidrule(lr){15-17} \cmidrule(lr){20-22} \cmidrule(lr){24-26}
\texttt{id} & \texttt{mean} & \texttt{\#b} & \texttt{time} &  & \texttt{mean} & \texttt{\#b} & \texttt{time} & \texttt{best} &  & \texttt{mean} & \texttt{\#b} & \texttt{time} &  & \texttt{mean} & \texttt{\#b} & \texttt{time} & \texttt{best} &  & \texttt{mean} & \texttt{\#b} & \texttt{time} &  & \texttt{mean} & \texttt{\#b} & \texttt{time} & \texttt{best} \\
\midrule
0 & \cellcolor{gray!20} \textbf{5.0575160} & \textbf{40}\rmark{\textcolor[HTML]{808080}{\faArrowUp}} & 0.87 &  & \cellcolor{gray!20} \textbf{5.0575160} & \textbf{40}\rmark{\textcolor[HTML]{808080}{\faArrowUp}} & \textbf{0.04}\rmark{\textcolor[HTML]{4C78A8}{\faBolt}} & 5.0575160 &  & 7.0156623 & 39 & 5.60 &  & \cellcolor[HTML]{B6CAE3} \textbf{7.0156453} & \textbf{40}\rmark{\textcolor[HTML]{4C78A8}{\faArrowUp}} & \textbf{1.41}\rmark{\textcolor[HTML]{4C78A8}{\faBolt}} & 7.0156453 &  & \cellcolor[HTML]{F0B6B6} \textbf{9.6706542} & \textbf{2}\rmark{\textcolor[HTML]{D95F5F}{\faArrowUp}} & 92.60 &  & 9.6904399 & 0 & \textbf{87.50}\rmark{\textcolor[HTML]{4C78A8}{\faBolt}} & 9.6328396 \\
\end{tabular}
}
\end{table}

Several consistent trends can be observed. In terms of mean solution quality, \textsc{RL-CMSA} generally outperforms HGA. For the smallest instances ($n=50$), the results show that \textsc{RL-CMSA} and \textsc{HGA} tie for small values of $m$, with both methods producing a best-found solution in all executions. As $m$ increases, however, \textsc{RL-CMSA} begins to pull ahead on some instances: while \textsc{HGA} no longer attains a best-found solution consistently across runs, \textsc{RL-CMSA} still does, yielding better average objective values. For larger instances ($n=100$ and $n=200$), \textsc{RL-CMSA} more regularly achieves better average objective values than HGA; the main exception occurs for $m=1\%$ (see Table~\ref{tab:summary_m_1}(a)), where \textsc{HGA} seems to outperform \textsc{RL-CMSA} on the largest instances ($n=200$). Note that the number of runs producing a best-found solution (\texttt{\#b}) is an indication of algorithm robustness. For $n=50$ and $n=100$, \textsc{RL-CMSA} reaches a best-found solution for almost all instances and in nearly all of the $40$ executions, indicating very stable performance. For $n=200$, \textsc{RL-CMSA} initially struggles to match the best solution frequently (particularly for small $m$), but this behaviour greatly improves as $m$ increases: the \texttt{\#b} counts grow with $m$, and for $m=15\%$ \textsc{RL-CMSA} again reaches the best solution in almost all runs for most instances.

Finally, regarding running times, \textsc{RL-CMSA} is consistently faster for $n=50$ and $n=100$, reaching its best obtained solutions within noticeably lower CPU times than HGA. For $n=200$, the comparison depends on $m$: for $m=1\%$ \textsc{HGA} is generally faster, for $m \in \{5\% ,10\%\}$ both methods exhibit similar running times (with instance-dependent differences), and for $m=15\%$ \textsc{RL-CMSA} becomes faster in most instances. Overall, these results show that \textsc{RL-CMSA} dominates in average objective value and best-run frequency for almost all settings, with the main drawback appearing for the largest instances when $m$ is small. With increasing $m$, \textsc{RL-CMSA} improves both robustness and runtime on large instances and becomes the preferred option.

The weaker performance of \textsc{RL-CMSA} for small values of $m$ can be explained by the intrinsic mechanics of the method. As $m$ increases, tours become shorter, which makes the \emph{Solve} step more effective: the sub-instance contains shorter route components that can be recombined in many feasible ways, increasing the number of promising route combinations and thus facilitating the construction of high-quality solutions. In other words, a given sub-instance offers greater combinatorial flexibility when routes are short. In contrast, when $m$ is very small, routes are necessarily much longer and substantially harder to combine. Consequently, the \emph{Solve} step provides less benefit. This also explains why the tuning process tends to favour larger sub-instances in this setting: retaining a bigger pool of routes increases the chance of including routes that can be combined into a competitive solution.

\begin{table}[H]
\centering
\caption{Statistical tests comparing \textsc{RL-CMSA} vs HGA, paired by instance.}
\label{tab:wilcoxon_grid_20}
\resizebox{\columnwidth}{!}{%
\begin{tabular}{rcccccc}
\toprule
 \multicolumn{1}{r}{\texttt{n}} & \multicolumn{2}{c}{$50$} & \multicolumn{2}{c}{$100$} & \multicolumn{2}{c}{$200$} \\
\cmidrule(lr){2-3} \cmidrule(lr){4-5} \cmidrule(lr){6-7}
\texttt{m}  & $r_{rb}$ & $p$-value & $r_{rb}$ & $p$-value & $r_{rb}$ & $p$-value \\
\midrule
$1\%$  & 0.0000 & 1.000e+00 & -1.0000 & \textbf{4.883e-04}\psigC & 0.3714 & 5.000e-01 \\

$5\%$  & -1.0000 & 1.000e+00 & -1.0000 & \textbf{4.196e-05}\psigC & -1.0000 & \textbf{2.289e-05}\psigC \\

$10\%$  & -1.0000 & 5.000e-01 & -1.0000 & \textbf{2.747e-04}\psigC & -0.8762 & \textbf{1.175e-03}\psigB \\

$15\%$  & -1.0000 & \textbf{3.906e-02}\psigA & -0.8242 & \textbf{3.662e-02}\psigA & -1.0000 & \textbf{4.196e-05}\psigC \\

\bottomrule
\end{tabular}
}
\end{table}

Table~\ref{tab:wilcoxon_grid_20} summarises the paired statistical comparison between \textsc{RL-CMSA} and \textsc{HGA} for each combination of $m \in \{1\%, 5\%, 10\%, 15\%\}$ and problem size $n \in \{50,100,200\}$. For each instance, we compute the mean objective value over the $40$ independent runs of each algorithm and then apply a paired Wilcoxon signed-rank test across instances. The table reports the rank-biserial correlation $r_{rb}$ as an effect-size measure and the corresponding Holm-corrected $p$-values. Negative values of $r_{rb}$ indicate that \textsc{RL-CMSA} tends to obtain better performance than \textsc{HGA}, and viceversa.

These statistical results confirm that \textsc{RL-CMSA} is statistically superior in most settings. For $n=100$, \textsc{RL-CMSA} significantly outperforms \textsc{HGA} for all considered values of $m$, with very large effect sizes ($r_{rb}=-1$) and corrected $p$-values below $0.05$. For $n=200$, the advantage is also significant for $m \in \{5\%, 10\%, 15\%\}$, again with large negative effect sizes, whereas for $m=1\%$ the difference is not significant and the effect size favours \textsc{HGA}. For $n=50$, a significant difference is only observed for $m=15\%$, whereas for $m \in \{1\%, 5\%, 10\%\}$ the comparison is not significant after correction. This indicates that, on the smallest instances and for smaller values of $m$, \textsc{HGA} and \textsc{RL-CMSA} behave similarly.

\begin{figure}[h!]
  \centering
  {\includegraphics[width=\columnwidth]{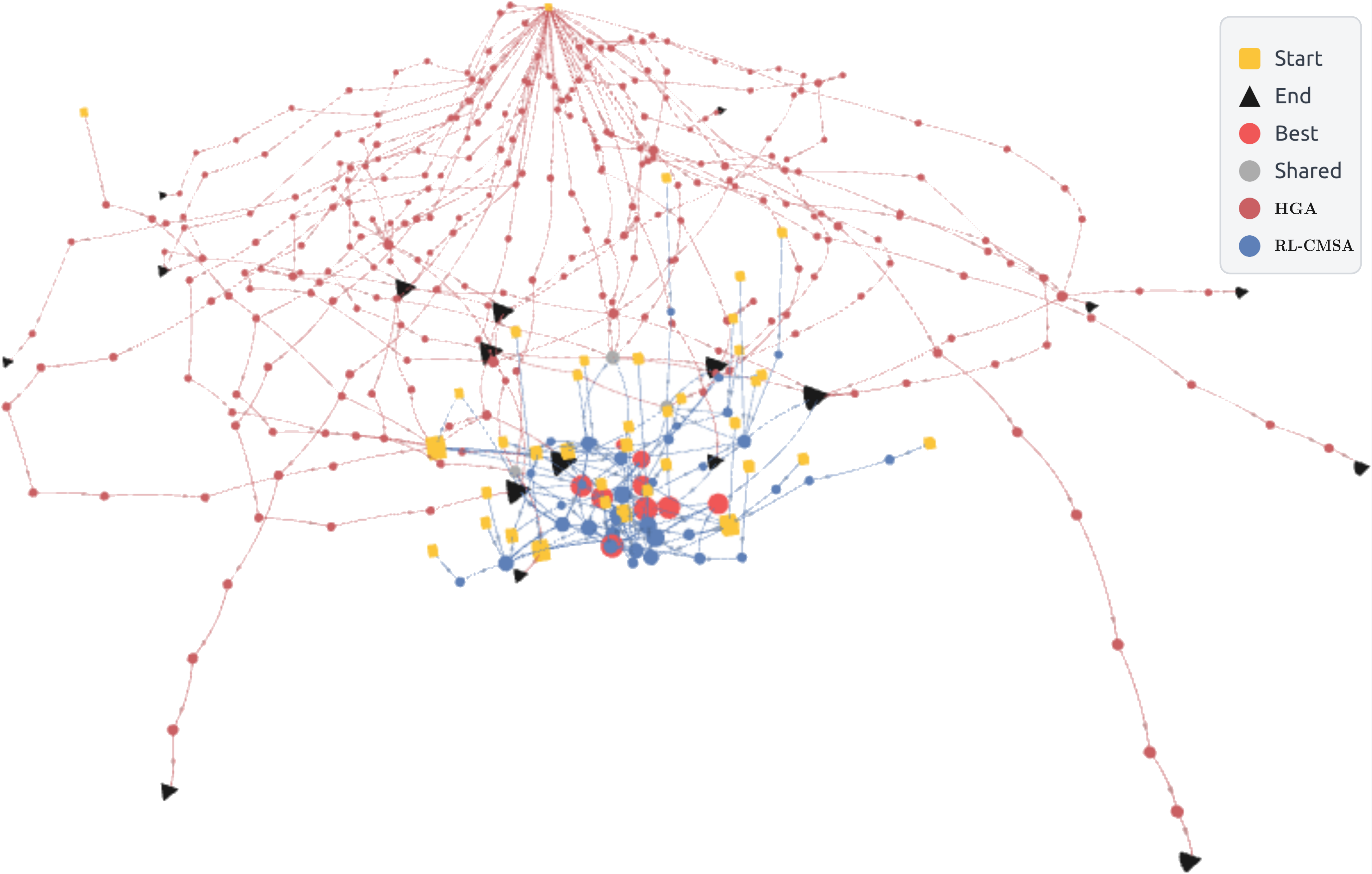}}
  \caption{STN graphic regarding instance 17 with $n=200$ and $m=10\%$.}
  \label{fig:stn}
\end{figure}

To better understand why \textsc{RL-CMSA} tends to produce more consistent results than HGA, we analyze a Search Trajectory Network (STN)~\cite{DBLP:journals/simpa/SartoriBO23} built from all executions of both algorithms on instance \texttt{id}=17 with $n=200$ and $m=10\%$ (a large instance with a medium-high number of vehicles). Figure~\ref{fig:stn} depicts the trajectories followed by the two methods during the search. Each node corresponds to one ore more solutions visited during an execution, and each directed edge indicates a transition to better solutions. Node colours follow the same scheme as in Table~\ref{tab:summary_m_1}, while grey nodes represent solutions visited by both algorithms. Yellow squares indicate initial solutions, and black triangles represent the final solutions of each run. The best solutions found are highlighted with a red circle; note that, as in the mTSP, it is very common that multiple different solutions share the same best objective value. The vertical position of a node encodes solution quality (fitness): higher nodes correspond to worse solutions, whereas lower nodes correspond to better ones. Since the STN is displayed with perspective, some black triangles may appear higher than the red-circled nodes depending on the viewpoint, although the red nodes correspond to the best solutions found. It can be observed that \textsc{HGA} typically starts from a narrow set of similar initial solutions and then disperses over a broader portion of the search space; several runs approach the region of high-quality solutions, but they rarely reach the best solutions. In contrast, \textsc{RL-CMSA} starts from more diverse initial solutions (often of better quality) and, in most runs, rapidly converges towards the same region where the best-found solutions are located. This behaviour suggests that \textsc{RL-CMSA} is more effective at guiding the search towards promising search space areas, which in turn explains its higher robustness across runs.

\begin{figure}[h!]
  \centering
  {\includegraphics[width=\columnwidth]{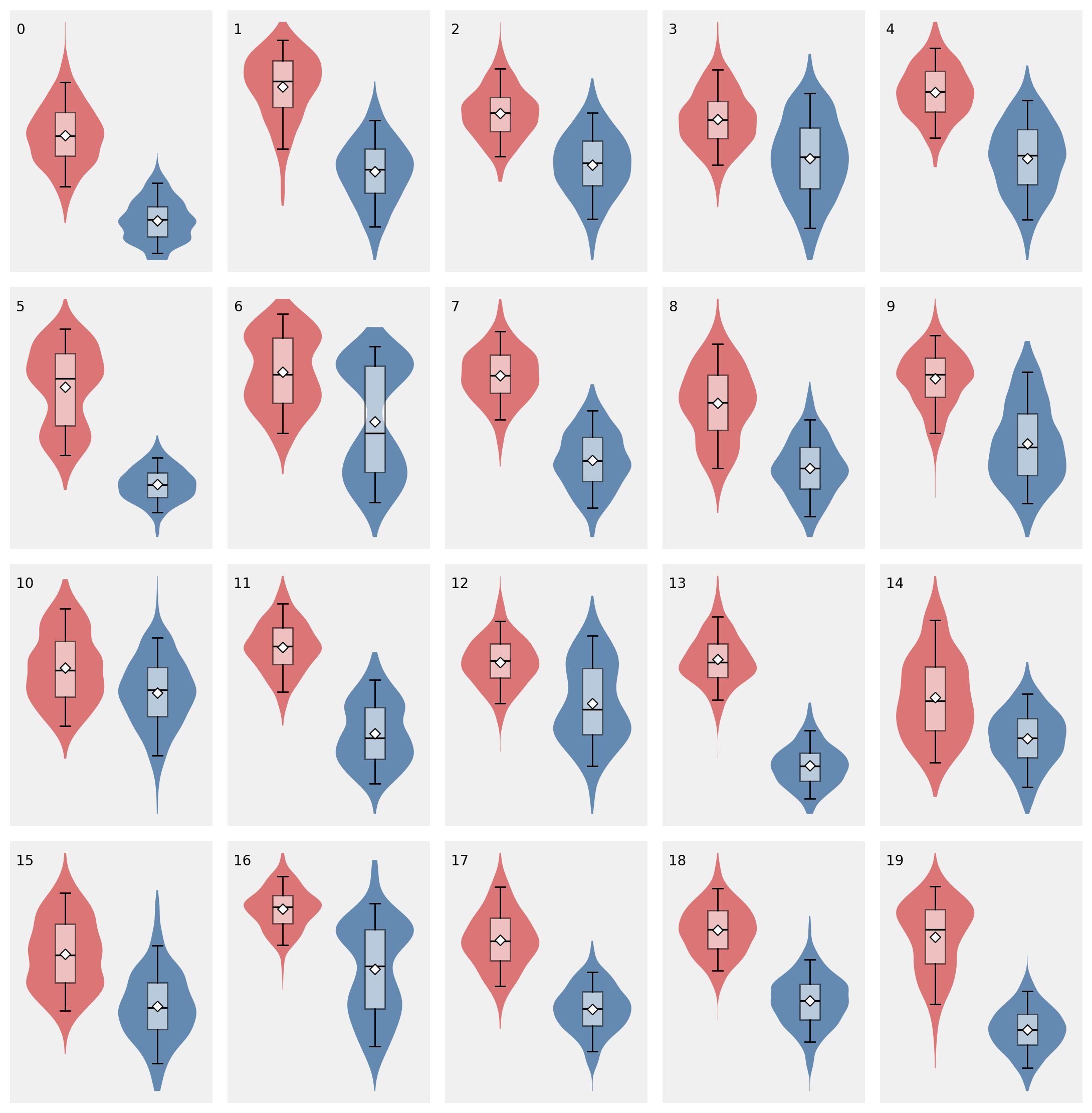}}
  \caption{Structural distances between the 40 best-found solutions per run (and per algorithm) for the 20 problem instances with $n=200$ and $m=10\%$.}
  \label{fig:diversity}
\end{figure}

Figure~\ref{fig:diversity} shows the distribution of the pairwise structural distances between the 40 solutions obtained by each algorithm on each of the 20 problem instances (with $n=200$ and $m=10\%$) using violinplots. Again, \textsc{HGA} is colored red, while \textsc{RL-CMSA} is colored blue. Wider regions of a violin indicate that many distances fall in that range, while the overlaid box/whiskers summarize the central tendency and spread (median and interquartile range), helping compare how dispersed the obtained solutions are for each method on each instance. Overall, a clear pattern emerges: \textsc{HGA} tends to produce solutions that are more different to each other (the red violins are usually shifted upward and often wider), which indicates higher diversity / more scattered search trajectories. In contrast, \textsc{RL-CMSA} typically produces solutions more similar among each other with tighter distributions (blue violins are generally lower and narrower). This is consistent with what was observed in the context of the STN graphic of Figure~\ref{fig:stn}.


\begin{table}[H]
\centering
\caption{Summary of the obtained results for the TSPLIB instances.}
\label{tab:tsplib_summary}
\resizebox{\textwidth}{!}{%
\begin{tabular}{r crccr@{\hspace{20pt}}c c crccr@{\hspace{20pt}}c c crccr@{\hspace{20pt}}c c crccr@{\hspace{20pt}}c}
\toprule
& \multicolumn{6}{c}{\texttt{eil51}} &  & \multicolumn{6}{c}{\texttt{berlin52}} &  & \multicolumn{6}{c}{\texttt{eil76}} &  & \multicolumn{6}{c}{\texttt{rat99}} \\

\cmidrule(lr){2-7} \cmidrule(lr){9-14} \cmidrule(lr){16-21} \cmidrule(lr){23-28}

\textit{} & \multicolumn{2}{c}{\textcolor[HTML]{D95F5F}{{\bfseries\scshape HGA}}} & \multicolumn{1}{c}{} & \multicolumn{2}{c}{\textcolor[HTML]{4C78A8}{{\bfseries\scshape RL-CMSA}}} & \multicolumn{1}{c}{} &  &
  \multicolumn{2}{c}{\textcolor[HTML]{D95F5F}{{\bfseries\scshape HGA}}} & \multicolumn{1}{c}{} & \multicolumn{2}{c}{\textcolor[HTML]{4C78A8}{{\bfseries\scshape RL-CMSA}}} & \multicolumn{1}{c}{} &  &
  \multicolumn{2}{c}{\textcolor[HTML]{D95F5F}{{\bfseries\scshape HGA}}} & \multicolumn{1}{c}{} & \multicolumn{2}{c}{\textcolor[HTML]{4C78A8}{{\bfseries\scshape RL-CMSA}}} & \multicolumn{1}{c}{} &  &
  \multicolumn{2}{c}{\textcolor[HTML]{D95F5F}{{\bfseries\scshape HGA}}} & \multicolumn{1}{c}{} & \multicolumn{2}{c}{\textcolor[HTML]{4C78A8}{{\bfseries\scshape RL-CMSA}}}  \\
  
\cmidrule(lr){2-3} \cmidrule(lr){5-6} \cmidrule(lr){9-10} \cmidrule(lr){12-13} \cmidrule(lr){16-17} \cmidrule(lr){19-20} \cmidrule(lr){23-24} \cmidrule(lr){26-27}
\texttt{m}
& \texttt{mean} & \texttt{\#b} &  & \texttt{mean} & \texttt{\#b} & \texttt{best} &  &
  \texttt{mean} & \texttt{\#b} &  & \texttt{mean} & \texttt{\#b} & \texttt{best} &  &
  \texttt{mean} & \texttt{\#b} &  & \texttt{mean} & \texttt{\#b} & \texttt{best} &  &
  \texttt{mean} & \texttt{\#b} &  & \texttt{mean} & \texttt{\#b} & \texttt{best} \\
\midrule

$1\%$
& 222.75794 & 37 &  & \cellcolor[HTML]{B6CAE3} \textbf{222.73337} & \textbf{40}\rmark{\textcolor[HTML]{4C78A8}{\faArrowUp}} & 222.73337 &  &
  \cellcolor{gray!20} \textbf{4110.21310} & \textbf{40}\rmark{\textcolor[HTML]{808080}{\faArrowUp}} &  & \cellcolor{gray!20} \textbf{4110.21310} & \textbf{40}\rmark{\textcolor[HTML]{808080}{\faArrowUp}} & 4110.21310 &  &
  281.22379 & 25 &  & \cellcolor[HTML]{B6CAE3} \textbf{280.85394} & \textbf{40}\rmark{\textcolor[HTML]{4C78A8}{\faArrowUp}} & 280.85394 &  &
  \cellcolor{gray!20} \textbf{665.99090} & \textbf{40}\rmark{\textcolor[HTML]{808080}{\faArrowUp}} &  & \cellcolor{gray!20} \textbf{665.99090} & \textbf{40}\rmark{\textcolor[HTML]{808080}{\faArrowUp}} & 665.99090 \\

$5\%$
& \cellcolor{gray!20} \textbf{159.57151} & \textbf{40}\rmark{\textcolor[HTML]{808080}{\faArrowUp}} &  & \cellcolor{gray!20} \textbf{159.57151} & \textbf{40}\rmark{\textcolor[HTML]{808080}{\faArrowUp}} & 159.57151 &  &
  \cellcolor{gray!20} \textbf{3069.58569} & \textbf{40}\rmark{\textcolor[HTML]{808080}{\faArrowUp}} &  & \cellcolor{gray!20} \textbf{3069.58569} & \textbf{40}\rmark{\textcolor[HTML]{808080}{\faArrowUp}} & 3069.58569 &  &
  \cellcolor{gray!20} \textbf{159.50091} & \textbf{40}\rmark{\textcolor[HTML]{808080}{\faArrowUp}} &  & \cellcolor{gray!20} \textbf{159.50091} & \textbf{40}\rmark{\textcolor[HTML]{808080}{\faArrowUp}} & 159.50091 &  &
  450.33361 & 0 &  & \cellcolor[HTML]{B6CAE3} \textbf{450.29462} & \textbf{24}\rmark{\textcolor[HTML]{4C78A8}{\faArrowUp}} & 450.27258 \\

$10\%$
& \cellcolor{gray!20} \textbf{118.13375} & \textbf{40}\rmark{\textcolor[HTML]{808080}{\faArrowUp}} &  & \cellcolor{gray!20} \textbf{118.13375} & \textbf{40}\rmark{\textcolor[HTML]{808080}{\faArrowUp}} & 118.13375 &  &
  \cellcolor{gray!20} \textbf{2440.92196} & \textbf{40}\rmark{\textcolor[HTML]{808080}{\faArrowUp}} &  & \cellcolor{gray!20} \textbf{2440.92196} & \textbf{40}\rmark{\textcolor[HTML]{808080}{\faArrowUp}} & 2440.92196 &  &
  127.56176 & 39 &  & \cellcolor[HTML]{B6CAE3} \textbf{127.56175} & \textbf{40}\rmark{\textcolor[HTML]{4C78A8}{\faArrowUp}} & 127.56175 &  &
  \cellcolor{gray!20} \textbf{436.44014} & \textbf{40}\rmark{\textcolor[HTML]{808080}{\faArrowUp}} &  & \cellcolor{gray!20} \textbf{436.44014} & \textbf{40}\rmark{\textcolor[HTML]{808080}{\faArrowUp}} & 436.44014 \\

$15\%$
& \cellcolor{gray!20} \textbf{112.07141} & \textbf{40}\rmark{\textcolor[HTML]{808080}{\faArrowUp}} &  & \cellcolor{gray!20} \textbf{112.07141} & \textbf{40}\rmark{\textcolor[HTML]{808080}{\faArrowUp}} & 112.07141 &  &
  \cellcolor{gray!20} \textbf{2440.92196} & \textbf{40}\rmark{\textcolor[HTML]{808080}{\faArrowUp}} &  & \cellcolor{gray!20} \textbf{2440.92196} & \textbf{40}\rmark{\textcolor[HTML]{808080}{\faArrowUp}} & 2440.92196 &  &
  \cellcolor{gray!20} \textbf{127.56175} & \textbf{40}\rmark{\textcolor[HTML]{808080}{\faArrowUp}} &  & \cellcolor{gray!20} \textbf{127.56175} & \textbf{40}\rmark{\textcolor[HTML]{808080}{\faArrowUp}} & 127.56175 &  &
  \cellcolor{gray!20} \textbf{436.44014} & \textbf{40}\rmark{\textcolor[HTML]{808080}{\faArrowUp}} &  & \cellcolor{gray!20} \textbf{436.44014} & \textbf{40}\rmark{\textcolor[HTML]{808080}{\faArrowUp}} & 436.44014 \\
\bottomrule
\end{tabular}
}
\end{table}

Finally, Table~\ref{tab:tsplib_summary} reports results on the \texttt{TSPLIB} instances using the same summary format as before. Here, the four horizontal blocks correspond to the tested instances (ordered by increasing size), and rows to the different $m$ values.
Overall, \textsc{RL-CMSA} matches \textsc{HGA}’s mean objective in most settings and improves it in five of them. It achieves the best-known solution in all executions except for one case (\texttt{rat99} with $m=5\%$), where it still reaches the best in more than half of the runs. The table omits the time to reach the best-found solutions due to space limitations. Nonetheless, \textsc{RL-CMSA} is generally faster than \textsc{HGA}. The few exceptions coincide with settings where \textsc{RL-CMSA} is also stronger in solution quality, suggesting that additional search time can help it escape local optima and converge to better solutions.

\section{Conclusions and Future Work}

In this paper, we introduced \textsc{RL-CMSA}, a hybrid \emph{Construct--Merge--Solve--Adapt} framework with reinforcement-guided clustering for the min--max mTSP. The method generates diverse candidate solutions, maintains a compact pool of high-quality routes, and repeatedly solves a restricted set-covering MILP to recombine them, while updating pairwise $q$-values to bias future constructions. On randomly generated instances, \textsc{RL-CMSA} is robust and generally improves both the mean objective value and the frequency of best runs compared to the state-of-the-art \textsc{HGA}. These gains are statistically significant in most settings, and become more pronounced as the number of cities and vehicles increases. The main exception occurs on the largest instances when the number of routes is very small. Moreover, on the considered \texttt{TSPLIB} instances, \textsc{RL-CMSA} outperforms \textsc{HGA} in five settings. In the remaining cases, the two methods achieve comparable results, although \textsc{RL-CMSA} is generally faster. An analysis of the search dynamics further indicates that \textsc{RL-CMSA} tends to drive runs toward a single high-quality region, yielding more consistent outcomes, whereas \textsc{HGA} explores more broadly and exhibits greater run-to-run variability.

As future work, we will strengthen \textsc{RL-CMSA} by enriching the route pool, e.g., by integrating additional large-scale neighborhoods, and by extending the reinforcement scheme to learn higher-order route features beyond pairwise co-occurrences. We also plan to evaluate the approach on broader and larger benchmark families and to extend it to more general routing settings with additional constraints.

\section*{Acknowledgements}
This work was supported in part by the Department of Research and Universities of the Government of Catalonia by means of an European Social Fund (ESF)-Founded Pre-Doctoral Grant of the Catalan Agency for Management of University and Research Grants (AGAUR) under Grant 2022 FI\_B 00903, in part by the Agencia Estatal de Investigación (AEI) under grants PID-2020-112581GB-C21 (MOTION) and PID2022-136787NB-I00 (ACISUD), and in part by the European Commission–NextGenerationEU, through Momentum CSIC Programme: Develop Your Digital Talent, under Project MMT24-IIIA-02.

\bibliographystyle{plain} 
\bibliography{sample-base} 



\end{document}